\documentclass{article}

\usepackage{microtype}
\usepackage{graphicx}
\usepackage{subcaption}
\usepackage{booktabs} 

\usepackage{hyperref}

\usepackage{xspace}
\usepackage{color, colortbl}

\usepackage{multirow}
\usepackage{makecell}

\newcommand{\stack}[2]{\makecell{#1\\#2}}
\newcommand{\fst}[1]{\textbf{#1}}
\newcommand{\snd}[1]{\underline{#1}}

\newcommand{\secref}[1]{\S\ref{#1}}

\usepackage[preprint]{icml2026}

\usepackage{amsmath}
\usepackage{amssymb}
\usepackage{mathtools}
\usepackage{amsthm}

\usepackage[capitalize,noabbrev]{cleveref}

\theoremstyle{plain}

\theoremstyle{definition}

\theoremstyle{remark}

\usepackage[textsize=tiny]{todonotes}

\icmltitlerunning{VP-VAE: Rethinking Vector Quantization via Adaptive Vector Perturbation}

\begin{document}

    \twocolumn[
        \icmltitle{VP-VAE: Rethinking Vector Quantization via Adaptive Vector Perturbation}



        \icmlsetsymbol{equal}{*}

        \begin{icmlauthorlist}
            \icmlauthor{Linwei Zhai}{sch}
            \icmlauthor{Han Ding}{sch}
            \icmlauthor{Mingzhi Lin}{sch}
            \icmlauthor{Cui Zhao}{sch}
            \icmlauthor{Fei Wang}{sch}
            \icmlauthor{Ge Wang}{sch}
            \icmlauthor{Zhi Wang}{sch}
            \icmlauthor{Wei Xi}{sch}
        \end{icmlauthorlist}

        \icmlaffiliation{sch}{Xi'an Jiaotong University, China}

        \icmlcorrespondingauthor{Han Ding}{dinghan@xjtu.edu.cn}

        \icmlkeywords{Machine Learning, Generative Models, Vector Quantization, VQ-VAE}

        \vskip 0.3in
    ]



    \printAffiliationsAndNotice{}  

    \begin{abstract}
        Vector Quantized Variational Autoencoders (VQ-VAEs) are fundamental to modern generative modeling, yet they often suffer from training instability and ``codebook collapse'' due to the inherent coupling of representation learning and discrete codebook optimization.
        In this paper, we propose \textbf{VP-VAE} (Vector Perturbation VAE), a novel paradigm that decouples representation learning from discretization by eliminating the need for an explicit codebook during training.
        Our key insight is that, from the neural network's viewpoint, performing quantization primarily manifests as injecting a structured perturbation in latent space.
        Accordingly, VP-VAE replaces the non-differentiable quantizer with \emph{distribution-consistent} and \emph{scale-adaptive} latent perturbations generated via Metropolis--Hastings sampling.
        This design enables stable training without a codebook while making the model robust to inference-time quantization error.
        Moreover, under the assumption of approximately uniform latent variables, we derive \textbf{FSP} (Finite Scalar Perturbation), a lightweight variant of VP-VAE that provides a unified theoretical explanation and a practical improvement for FSQ-style fixed quantizers.
        Extensive experiments on image and audio benchmarks demonstrate that VP-VAE and FSP improve reconstruction fidelity and achieve substantially more balanced token usage, while avoiding the instability inherent to coupled codebook training.
        {\renewcommand{\thefootnote}{\fnsymbol{footnote}}
        \footnote{The example code is available at \url{https://github.com/zhai-lw/vp-vae}. }}
    \end{abstract}

    \section{Introduction}\label{sec:intro}

Discrete tokenization of continuous signals is a core element in modern generative modeling.
By converting high-dimensional data (e.g., images or audio) into compact discrete sequences, one can directly leverage sequence models such as Transformers~\cite{chen2025, esser2021, rombach2022} for downstream tasks, including image/audio generation and text-to-image/audio synthesis~\cite{DALL-E2021, MaskGIT2022, LFQ2024, AudioLM2023, brendel2024}.
Vector-Quantized VAEs (VQ-VAEs)~\cite{VQVAE2017} and their variants~\cite{yu2021,SoundStream2021, RandomVQ2022, FSQ2023, huh2023, fifty2024, SimVQ2025} remain a dominant approach for learning such discrete representations.

However, optimizing VQ-VAEs is often unstable due to an inherent \emph{coupling} between representation learning and codebook learning.
During training, encoder outputs must be assigned to their nearest codes via a non-differentiable \(\arg\min\) operator, while the codebook vectors must simultaneously move to fit the evolving distribution of encoder outputs.
To enable gradient-based optimization, surrogate gradient estimators—most commonly the straight-through estimator (STE)~\cite{STE2013X}—are widely adopted.
While STE enables backpropagation, it introduces a gradient mismatch: the forward pass applies a non-differentiable discretization, whereas the backward pass relies on a continuous straight-through approximation.
This gradient mismatch, together with non-stationary optimization targets induced by concurrent codebook updates, makes training sensitive and difficult to scale.
A prominent failure mode is \emph{codebook collapse} (or dead codes): only a small subset of codes are repeatedly selected, while many codes receive few or no assignments and thus stop updating.
Importantly, this forms a self-reinforcing vicious cycle: early imbalances in code selection reduce updates to rarely chosen codes, which in turn makes them even less likely to be selected later, effectively shrinking the model's usable discrete capacity and reducing the effective bitrate.

Prior work broadly follows two directions, each addressing the above issues only partially.
\begin{enumerate}
    \item \textit{Improved coupled codebook learning.}
    A line of methods improves the joint optimization of encoder/decoder and codebook via additional regularization or training heuristics (e.g., code resets, orthogonality constraints, or reparameterizations)~\cite{yu2021, DALL-E2021, SoundStream2021, SQ-VAE2022, huh2023, zheng2023, zhang2023, fifty2024, SimVQ2025}.
    While these methods are sometimes effective, they introduce extra objectives and hyperparameters, and the underlying coupling remains.

    \item \textit{Predefined codebooks.}
    Another line removes codebook learning by adopting fixed quantizers, such as lattice-based scalar quantization schemes (e.g., FSQ, LFQ) or predefined Gaussian grids~\cite{FSQ2023, LQ2023, LFQ2024, BSQ2024, TokenBridge2025}.
    These approaches are typically stable, yet they impose a rigid prior on the latent space; when the learned latent distribution deviates from the assumed grid geometry, the quantization capacity can be inefficiently used (e.g., many tokens carry redundant probability mass while other regions remain under-represented).
\end{enumerate}

This paper asks a fundamental question: \emph{Can we train the encoder--decoder without knowing the codebook at all, and only instantiate the codebook after the continuous representation has converged?}
Our key observation is that, for the decoder, quantization is effectively experienced as a bounded and local perturbation to the latent representation---namely the \emph{quantization error}.
Therefore, instead of explicitly performing discrete quantization during training (and dealing with its optimization difficulties), we propose to train the decoder to be robust to \emph{appropriate} latent perturbations that emulate the effect of quantization.
If the perturbation distribution matches the magnitude and locality of the quantization error incurred at inference time, the model can be trained without an explicit codebook, and the codebook can be generated after convergence.

To instantiate this idea, we propose \textbf{VP-VAE} (Vector Perturbation VAE).
VP-VAE replaces the discrete quantization operator in training with an \emph{adaptive vector perturbation} mechanism.
The key challenge is to design perturbations that serve as a good approximation for quantization error: a naive choice such as isotropic Gaussian noise is insufficient, because it alters the latent distribution and frequently pushes samples into low-density regions, forcing the decoder to allocate capacity to reconstruct from latent vectors that are unlikely to occur under the data distribution.
In contrast, VP-VAE generates perturbations via a Metropolis--Hastings (MH) sampling procedure with non-parametric density estimation, so that the perturbed latent vectors (denoted by \(\tilde{z}\)) remain in high-probability regions of the original latent space (denoted by \(z\)).
At the same time, the perturbation magnitude is designed to be aligned with the expected quantization error implied by a target codebook size.
In experiments, this decoupled training paradigm yields improved reconstruction quality and substantially more balanced token usage.

Moreover, under a simplifying assumption that each latent dimension is constrained to an (approximately) uniform distribution over a bounded interval, our general framework naturally simplifies to a lightweight scalar variant, namely \textbf{FSP} (Finite Scalar Perturbation).
We show that FSP aligns with the principles of optimal scalar quantization in the Lloyd--Max sense, and outperforms strong fixed-quantizer baselines such as FSQ~\cite{FSQ2023, noise_FSQ2024, symmetry_FSQ2024}.
Instead of quantizing to grid edges, FSP corresponds to training with centered perturbations and decoding with interval centroids, providing both a solid theoretical foundation and superior experimental performance.

Our contributions are summarized as follows:

    $\bullet$ We propose VP-VAE, a new training paradigm that decouples representation learning from codebook learning, effectively preventing codebook collapse and enabling a more stable, theoretically grounded, and flexible approach to learning discrete representations from continuous signals.
    
    $\bullet$ We develop a perturbation mechanism based on Metropolis--Hastings sampling that injects \emph{scale-adaptive} and \emph{distribution-consistent} latent noise, serving as an approximation for quantization error without requiring a learned codebook during training.
    
    $\bullet$ We derive FSP, a lightweight scalar variant that assumes an approximately uniform latent space, providing both theoretical insight into and empirical improvements over FSQ-style fixed quantizers.
    
    $\bullet$ Across image and audio benchmarks, VP-VAE and FSP consistently improve reconstruction fidelity while achieving higher and more balanced codebook utilization.
    \section{Related Work}\label{sec:related_work}

Existing approaches to neural discrete representation learning largely fall into two categories: methods that seek to mitigate the instability arising from joint model–codebook optimization, and methods that bypass this coupling by relying on fixed or predefined quantization targets.


\subsection{Coupled Codebook--Network Optimization}
The foundational VQ-VAE~\cite{VQVAE2017} introduced the concept of learning a discrete latent bottleneck via nearest-neighbor lookup, utilizing the Straight-Through Estimator (STE) to approximate gradients for the encoder.
While effective, this paradigm suffers inherently from the ``codebook collapse'' phenomenon, where a significant portion of code vectors receive no gradients and become inactive.
This occurs because the codebook update (moving codes toward encoder outputs) and the encoder update (mapping inputs to codes) are coupled in a feedback loop that rewards already-frequent codes, often resulting in low effective bitrate~\cite{SQ-VAE2022, FSQ2023}.

To mitigate this, substantial efforts have focused on regularization and heuristic stabilization.
Early works employed Exponential Moving Average (EMA) updates and ``restart'' heuristics---manually re-initializing dead codes with active encoder outputs---to maintain high codebook utilization~\cite{VQ-VAE22019, zheng2023}.
More recent approaches focus on refining gradient estimation and update dynamics.
For instance, SQ-VAE~\cite{SQ-VAE2022} replaces quantization with stochastic approximation via Gaussian smoothing to improve gradient estimation.
Similarly, SimVQ~\cite{SimVQ2025} argues that the collapse stems from disjoint optimization updates; it proposes reparameterizing the codebook via a learnable linear layer to ensure gradients flow to all codes simultaneously.

Despite these advancements, these methods fundamentally retain the coupled training dynamics: the encoder must still track a moving target (the codebook), and the codebook must adapt to a shifting encoder distribution.
This interdependence often necessitates complex extra losses (e.g., commitment, orthogonality, or entropy loss) and renders the training process sensitive to hyperparameter selection~\cite{MaskGIT2022, shin2023, SimVQ2025}.
Our proposed VP-VAE differs by removing this coupling entirely during the representation learning phase.

\subsection{Predefined Codebook Quantization}
Recognizing the drawbacks of joint optimization, a parallel line of research explores decoupling representation learning from codebook generation by imposing fixed geometric priors.
RandomVQ~\cite{RandomVQ2022} demonstrated that in self-supervised speech learning, a randomly initialized and frozen codebook can yield competitive representations, suggesting that the precise location of codes is often less critical than the discretization mechanism itself.

Building on this, Finite Scalar Quantization (FSQ)~\cite{FSQ2023} and LFQ~\cite{LFQ2024} eliminate the vector lookup entirely, instead projecting latent dimensions onto fixed scalar grids forming a hypercube.
By removing the learnable codebook, these methods achieve high stability and near-100\% codebook utilization.
However, they impose a rigid bias: they assume the encoder can reshape the latent distribution into a factorial uniform grid.
If the natural latent distribution of the data is non-uniform or complex, these rigid grids lead to inefficient space allocation.

Most recently, TokenBridge~\cite{TokenBridge2025} proposes discretizing the latent space of a pre-trained VAE.
While effective as a post-training strategy, it fundamentally relies on the \emph{predefined} structure imposed by the VAE's KL-regularization, which forces the data to approximate a standard normal distribution; this continuous space is then sliced into discrete tokens using fixed Gaussian percentiles.
Conceptually, this amounts to utilizing a \emph{predefined Gaussian grid}.
Like the scalar grid methods mentioned above, this approach shifts the burden of adaptation entirely onto the encoder, forcing the latent distribution to conform to a specific prior to adapt the quantizer.
In contrast, VP-VAE imposes no such global distributional constraints (e.g., Gaussianity).
Instead, our adaptive perturbation mechanism generates local perturbations that are consistent with the empirical latent density, training the decoder to be robust to quantization error without forcing the latent distribution to match a predefined grid.

    \section{Methodology}\label{sec:method}

This section introduces VP-VAE, a training framework that decouples representation learning from codebook learning by replacing discrete quantization with adaptive vector perturbation.
We elaborate on the model architecture, the perturbation-based training procedure, and the generation of an explicit codebook for inference (\secref{subsec:quant_to_perturb}).
We then present Finite Scalar Perturbation (FSP), a specialized variant of VP-VAE. When the latent distribution is approximately uniform, FSP enables a simplified perturbation and quantization scheme with reduced computational complexity (\secref{subsec:fsp}).

\subsection{Setup and Notation}\label{subsec:setup}

Let \(x\) denote an input sample, such as an image or an audio signal.
An encoder \(E\) maps \(x\) to a sequence of token features \(\{h_t\}_{t=1}^{T}\), where \(h_t \in \mathbb{R}^{C}\) and \(T\) is the number of tokens per sample.
In standard VQ-VAEs, quantization is performed directly in this \(C\)-dimensional embedding space (e.g., \(C = 128\) or higher).
Instead, in this paper, we propose a perturbation mechanism based on non-parametric density estimation, designed to ensure that perturbed latent vectors remain within high-density regions of the original latent distribution.
However, performing reliable density estimation in such high-dimensional spaces is non-trivial due to the curse of dimensionality~\cite{silverman1986density}.

To address this issue, VP-VAE introduces a \emph{low-dimensional quantization bottleneck}.
Specifically, each token feature is first projected into a lower-dimensional space via a learnable down-projection, and later mapped back to the original embedding space through a corresponding up-projection:
\begin{align}
    z = P_{\downarrow}(h_t) \in \mathbb{R}^{d}, \nonumber \\
    \tilde{h}_t = P_{\uparrow}(\tilde{z}) \in \mathbb{R}^{C}, \label{eq:proj}
\end{align}
where \(P_{\downarrow}: \mathbb{R}^{C} \to \mathbb{R}^{d}\) and \(P_{\uparrow}: \mathbb{R}^{d} \to \mathbb{R}^{C}\) denote the down- and up-projection operators, respectively.
Here, \(z\) and \(\tilde{z}\) represent the latent vectors before and after perturbation, as detailed in \secref{subsubsec:mh_perturb}.
Both perturbation (during training) and quantization (during testing) are performed in the compressed \(d\)-dimensional space (typically \(d \le 16\)), while the decoder reconstructs the input as \(\hat{x} = D(\{\tilde{h}_t\}_{t=1}^{T})\) from the up-projected features.
This bottleneck enables reliable density estimation while preserving sufficient representational capacity for high-fidelity reconstruction.

\subsection{VP-VAE: From Quantization to Perturbation}\label{subsec:quant_to_perturb}

In standard VQ, each latent vector \(z\) is mapped to its nearest code \(q(z) \in \mathcal{Q}\), introducing a \emph{quantization error} \(\epsilon = q(z) - z\).
Our key viewpoint is that, this operation can be viewed as injecting a structured perturbation into the latent space.
This viewpoint motivates a fundamentally different training strategy: instead of performing discrete quantization during training---which requires surrogate gradients and simultaneous codebook updates---we can train the decoder to be robust to latent perturbations that mimic the effect of quantization.

Following this point, VP-VAE replaces the discrete quantization operator with an explicit perturbation operator \(\mathcal{T}\):
\begin{align}
    \tilde{z} = \mathcal{T}(z; \mathcal{S}), \nonumber \\
    \hat{x} = D\big(\{P_{\uparrow}(\tilde{z})\}_{t=1}^{T}\big), \label{eq:perturb_operator}
\end{align}
where \(\mathcal{S}\) is a memory buffer containing recent latent vectors,  used to approximate the current latent distribution (detailed in \secref{subsubsec:scale_est}).
If the perturbation distribution accurately emulates quantization error, the model can learn to reconstruct from perturbed latent vectors without ever seeing an explicit codebook during training.

For this emulation to be effective, the perturbation operator \(\mathcal{T}\) must satisfy two critical requirements:
\begin{enumerate}
    \item \textbf{Scale alignment}: The perturbation magnitude should match the expected quantization error implied by a target codebook size \(K\).
    Perturbations that are too small may fail to prepare the decoder for inference-time quantization, while overly large perturbations unnecessarily degrade reconstruction quality.
    \item \textbf{Distribution consistency}: Perturbed latent vectors \(\tilde{z}\) should remain within high-density regions of the original latent distribution.
    Naive perturbation schemes, such as isotropic Gaussian noise, are ineffective because they often push samples into low-density or out-of-distribution regions, forcing the decoder to model unlikely latent configurations and wasting representational capacity.
\end{enumerate}

\subsubsection{Scale Alignment via Adaptive Radius Estimation}\label{subsubsec:scale_est}

To align the perturbation scale with the expected quantization error, we estimate how far a latent vector would typically move under nearest-neighbor quantization with a codebook of size \(K\).
Intuitively, if the latent space were partitioned into \(K\) Voronoi cells with roughly equal probability mass, each cell would contain approximately \(1/K\) of the latent distribution, and the quantization error would scale with the local cell radius.

We realize this intuition using a non-parametric approach.
We maintain a first-in-first-out (FIFO) queue \(\mathcal{S} = \{s_i\}_{i=1}^{|\mathcal{S}|}\) of recent latent vectors in \(\mathbb{R}^d\).
To mitigate redundancy caused by strong intra-sample token correlations, we update \(\mathcal{S}\) by randomly subsampling a small fraction of tokens from each minibatch rather than storing all tokens.

Given the queue \(\mathcal{S}\), let \(D_m(z | \mathcal{S})\) denote the Euclidean distance from \(z\) to its \(m\)-th nearest neighbor in \(\mathcal{S}\).
For a target codebook size \(K\), we define the \emph{local quantization radius} as:
\begin{equation}
    M = \left\lceil \frac{|\mathcal{S}|}{K} \right\rceil,
    \qquad
    R(z) = \eta \, D_M(z | \mathcal{S}),
    \label{eq:radius}
\end{equation}
where \(\eta > 0\) is a hyperparameter controlling the perturbation scale.
The rationale is as follows: if \(K\) codebook vectors were to partition the \(|\mathcal{S}|\) samples into regions of equal size, each region would contain approximately \(M = |\mathcal{S}|/K\) samples.
The distance to the \(M\)-th nearest neighbor thus serves as a data-adaptive indicator for the radius of the local Voronoi cell, automatically adjusting to both the local density and the target codebook size.

\subsubsection{Distribution Consistency via Metropolis--Hastings Sampling}\label{subsubsec:mh_perturb}

With the perturbation scale determined by Eq.~\eqref{eq:radius}, a naive approach would sample perturbations uniformly from a ball of radius \(R(z)\).
However, such isotropic noise ignores the geometry of the latent distribution, i.e., perturbations may push latent vectors into low-density regions or even outside the support of the learned latent distribution, creating a train--test mismatch.

To preserve distributional consistency, VP-VAE generates perturbations using a Metropolis--Hastings (MH) transition~\cite{metropolis1953equation, hastings1970monte} whose stationary distribution matches the empirical latent density.
A fundamental property of the MH algorithm is that the transition kernel leaves the target distribution \(\pi\) invariant: if \(z \sim \pi\), then \(\tilde{z} \sim \pi\) as well~\cite{robert1999monte}.
MH sampling naturally rejects proposals that move into low-density regions, ensuring that accepted perturbations keep \(\tilde{z}\) within high-probability regions.

\paragraph{Target density estimation.}
We approximate the latent density using a \(k\)-nearest-neighbor (kNN) estimator:
\begin{equation}
    \pi(z) \propto \frac{1}{\big(D_k(z | \mathcal{S})\big)^d},
    \label{eq:knn_density}
\end{equation}
where \(k\) is a small constant, \(D_k(z | \mathcal{S})\) is the distance to the \(k\)-th nearest neighbor, and \(d\) is the latent dimension.
For brevity, we write \(D_k(z) = D_k(z|\mathcal{S})\) and \(D_M(z) = D_M(z|\mathcal{S})\) when the context is clear.

\paragraph{Proposal distribution.}
Given a current latent vector \(z\), we propose a candidate \(z'\) by sampling uniformly from the \(d\)-dimensional ball centered at \(z\) with radius \(R(z)\):
\begin{equation}
    z' = z + u, \qquad u \sim \mathrm{Unif}\big(\mathcal{B}(0, R(z))\big).
    \label{eq:proposal}
\end{equation}
Uniform sampling in a ball can be implemented by sampling a random direction \(v \sim \mathcal{N}(0, I_d)\) and a random radius \(r = R(z) \cdot \rho^{1/d}\) with \(\rho \sim \mathrm{Unif}[0,1]\), then setting \(u = r \cdot v / \|v\|_2\).

\paragraph{Acceptance probability.}
The MH acceptance probability determines whether a proposed perturbation is applied, which is defined as:
\begin{equation}
    \alpha(z, z') = \min\left(1,\ \frac{\pi(z') \, g(z | z')}{\pi(z) \, g(z' | z)}\right),
    \label{eq:mh_accept}
\end{equation}
where \(g(\cdot | \cdot)\) denotes the proposal density.
Intuitively, this ratio compares how likely the proposed latent vector is under the target density relative to the current latent vector \(z\).
As a result, proposals that move toward higher-density regions are likely to be accepted, whereas moves into lower-density regions are rejected.

For the uniform-ball proposal, the reverse density \(g(z | z')\) is nonzero only if \(\|z - z'\| \le R(z')\).
Therefore, we explicitly enforce the support condition:
\begin{equation}
    \alpha(z, z') = 0 \quad \text{if } \|z - z'\| > R(z').
    \label{eq:support_check}
\end{equation}
When both support conditions are satisfied, substituting Eq.~\eqref{eq:knn_density} and the proposal densities yields:
\begin{equation}
    \alpha(z, z') = \min\left(1,\ \left(\frac{D_k(z) \cdot D_M(z)}{D_k(z') \cdot D_M(z')}\right)^{\!d}\right),
    \label{eq:mh_ratio}
\end{equation}
where we use \(R(\cdot) = \eta D_M(\cdot)\) from Eq.~\eqref{eq:radius}.

\paragraph{Perturbation operator.}
The final perturbed latent vector is obtained by applying the MH accept--reject rule:
\begin{equation}
    \tilde{z} =
    \begin{cases}
        z', & \text{with probability } \alpha(z, z'), \\
        z,  & \text{otherwise}.
    \end{cases}
    \label{eq:perturbed_latent}
\end{equation}
If the proposal is rejected, the latent vector passes through unperturbed.
The complete procedure is summarized in \cref{alg:vpvae_mh}.

\begin{algorithm}[t]
\caption{Vector Perturbation}
\label{alg:vpvae_mh}
\footnotesize
\begin{algorithmic}[1]
\REQUIRE Latent token \(z \in \mathbb{R}^d\), queue \(\mathcal{S}\), target codebook size \(K\), scale \(\eta\), kNN parameter \(k\).
\STATE \(M \leftarrow \lceil |\mathcal{S}| / K \rceil\)
\STATE Compute \(D_k(z), D_M(z)\) via kNN search on \(\mathcal{S}\)
\STATE \(R(z) \leftarrow \eta \cdot D_M(z)\)
\STATE Sample \(u \sim \mathrm{Unif}(\mathcal{B}(0, R(z)))\); \quad \(z' \leftarrow z + u\)
\STATE Compute \(D_k(z'), D_M(z')\) via kNN search on \(\mathcal{S}\)
\STATE \(R(z') \leftarrow \eta \cdot D_M(z')\)
\IF{\(\|z - z'\|_2 > R(z')\)}
\STATE \textbf{return} \(\tilde{z} \leftarrow z\)
\ENDIF
\STATE \(\text{ratio} \leftarrow \big(D_k(z) \cdot D_M(z) / (D_k(z') \cdot D_M(z'))\big)^d\)
\STATE \(\alpha \leftarrow \min(1, \text{ratio})\)
\STATE Sample \(b \sim \mathrm{Bernoulli}(\alpha)\)
\IF{\(b=1\)}
\STATE \textbf{return} \(\tilde{z}\leftarrow z'\)
\ELSE
\STATE \textbf{return} \(\tilde{z}\leftarrow z\)
\ENDIF
\end{algorithmic}
\end{algorithm}

\subsubsection{Optimization Objective}\label{subsubsec:objective}

VP-VAE is trained end-to-end with a reconstruction loss augmented by a latent normalization regularizer:
\begin{equation}
    \mathcal{L} = \mathcal{L}_{\text{rec}}(x, \hat{x}) + \mathcal{L}_{\text{norm}}.
    \label{eq:loss_total}
\end{equation}
The reconstruction loss \(\mathcal{L}_{\text{rec}}\) can be any differentiable metric appropriate for the data modality (e.g., MSE, perceptual loss, or a combination thereof).

The normalization regularizer encourages the latent distribution to be zero-mean and unit-variance along each dimension:
\begin{equation}
    \mathcal{L}_{\text{norm}} = \lambda_1 \|\mu_{batch}\|_2^2 + \lambda_2 \|\sigma_{batch}^2 - \mathbf{1}\|_2^2,
    \label{eq:loss_norm}
\end{equation}
where \(\mu_{batch}\) and \(\sigma_{batch}^2\) are the minibatch mean and variance of the \(d\)-dimensional latent vectors \(z\).
This regularizer serves two purposes: (i) it prevents latent collapse or explosion, and (ii) it facilitates scale estimation by ensuring that the queue \(\mathcal{S}\) contains samples from a well-behaved distribution.

\subsubsection{Codebook Generation}\label{subsubsec:codebook}

A distinguishing feature of VP-VAE is that no codebook exists during training.
Once training converges, we generate an explicit codebook offline by clustering the learned latent representations.
Specifically, we run the encoder over the training set to collect a large set of latent vectors \(z\) in \(\mathbb{R}^d\), then apply K-Means clustering (with K-Means++ initialization~\cite{arthur2006k}) to obtain \(K\) centroids, forming the codebook \(\mathcal{Q} = \{c_j\}_{j=1}^{K}\).

At inference time, each latent vector \(z\) is quantized via nearest-neighbor assignment:
\begin{equation}
    q(z) = \arg\min_{c \in \mathcal{Q}} \|z - c\|_2,
    \label{eq:inference_quant}
\end{equation}
followed by the up-projection \(P_{\uparrow}\) and decoding.
Because the encoder and decoder are trained to tolerate perturbations of magnitude \(R(z)\), and K-Means approximately minimizes quantization error, our model generalizes well to discrete quantization at inference time without requiring additional fine-tuning.

\subsection{Finite Scalar Perturbation (FSP)}\label{subsec:fsp}

The general VP-VAE framework handles arbitrary latent distributions via non-parametric density estimation.
However, when the latent distribution can be constrained to a simple and known form, the perturbation mechanism can be significantly simplified.
Under this observation, we further derive \textbf{FSP} (Finite Scalar Perturbation), a lightweight variant of VP-VAE that arises under the assumption of approximately uniform latent variables.

\paragraph{Uniform latent variables via ``CDF-like'' activation.}
Consider constraining each latent dimension to the unit interval \([0, 1]\) using a monotone ``CDF-like'' activation.
Let \(a \in \mathbb{R}^d\) denote pre-activation latent vectors, and define:
\begin{equation}
    z = g(a) \in [0, 1]^d,
    \label{eq:cdf_activation}
\end{equation}
where \(g\) is a smooth, monotonically increasing function (e.g., the sigmoid or a normal CDF).
In the idealized case where each \(a_i\) follows a distribution whose CDF equals \(g\), the probability integral transform implies that each \(z_i\) is marginally uniform on \([0,1]\).
In practice, we do not enforce exact matching of the full distribution; instead, we use a simple moment-matching regularizer on \(a\) (described below) and treat the resulting near-uniformity as an approximation.

Concretely, for FSP we apply a normalization regularizer to the pre-activation variables \(a\) rather than the post-activation variables \(z\):
\begin{equation}
    \mathcal{L}_{\text{norm}}^{\text{FSP}} = \lambda_1 \|\mu_{batch}(a)\|_2^2 + \lambda_2 \|\sigma_{batch}^2(a) - \sigma_g^2 \mathbf{1}\|_2^2,
    \label{eq:loss_norm_fsp}
\end{equation}
where \(\sigma_g^2\) is chosen based on the activation \(g\) so that \(z=g(a)\) is close to uniform on \([0,1]\) in practice.
For example, we use \(\sigma_g^2=1\) for the normal-CDF activation;
\(\sigma_g^2 \approx 0.8225\) for tanh activation (\(g(a)=(\tanh(a)+1)/2\));
and \(\sigma_g^2 \approx 3.29\) for sigmoid activation.

\paragraph{Lloyd--Max centroids.}
Under the uniform distribution, optimal scalar quantization is characterized by the Lloyd--Max conditions~\cite{max1960quantizing, lloyd1982least}.
For a uniform source on \([0, 1]\) with \(L\) quantization levels, the optimal reconstruction points correspond to the centroids of equal-width intervals:
\begin{equation}
    \mathcal{C} = \left\{\frac{\ell + 1/2}{L}\right\}_{\ell=0}^{L-1}.
    \label{eq:lloyd_centroids_fsp}
\end{equation}
FSP quantizes to these centroids rather than grid boundaries, which provides a principled improvement over rounding-based fixed grids (e.g., FSQ~\cite{FSQ2023}) when the (approximate) uniform-latent assumption holds.

\paragraph{Simplified perturbation mechanism.}
When the latent variables are (approximately) uniformly distributed on a bounded support, density estimation becomes trivial: the density is constant within the support and zero outside.
Consequently, the MH acceptance criterion reduces to a simple support check---any proposal that remains within \([0, 1]^d\) is accepted.
Moreover, the perturbation scale naturally aligns with the quantization bin width \(1/L\).

This yields a fully factorized, per-dimension perturbation rule.
For dimension \(i\) with \(L_i\) quantization levels, we propose:
\begin{equation}
    z'_i = z_i + u_i, \qquad u_i \sim \mathcal{U}\!\left(-\frac{\eta}{2L_i},\, \frac{\eta}{2L_i}\right),
    \label{eq:fsp_noise}
\end{equation}
and accept iff the proposal remains in the valid range:
\begin{equation}
    \tilde{z} =
    \begin{cases}
        z', & \text{if } z' \in [0, 1]^d, \\
        z,  & \text{otherwise}.
    \end{cases}
    \label{eq:fsp_reject}
\end{equation}

\paragraph{Training: perturb-or-quantize mixture.}
In practice, we find that mixing perturbation with explicit quantization during training improves gradient quality and accelerates convergence.
Inspired by noise-injection techniques for fixed quantizers~\cite{noise_FSQ2024}, FSP uses a stochastic mixture: at each forward pass, bounded perturbation (Eqs.~\eqref{eq:fsp_noise}--\eqref{eq:fsp_reject}) is applied with probability \(1/2\), while centroid quantization with STE is applied with probability \(1/2\):
\begin{equation}
    \tilde{z}_i = \frac{\ell_i + 1/2}{L_i}, \quad \ell_i = \mathrm{clip}\big(\lfloor L_i z_i \rfloor, 0, L_i - 1\big).
    \label{eq:fsp_quantize}
\end{equation}

In experiments, FSP consistently outperforms rounding-based fixed quantizers (e.g., FSQ) while remaining simpler than full VP-VAE.
However, when the latent distribution deviates from uniformity, VP-VAE retains a clear advantage due to its greater modeling flexibility.

    \begin{table*}[ht]
    \centering
    \caption{
        \textbf{In-domain reconstruction results.}
        Models are trained and evaluated on COCO (image) and LibriSpeech (audio).
        ``\texttt{-}'' indicates training failure due to severe codebook collapse.
        VP-VAE and FSP demonstrate consistent stability and high fidelity across both modalities, whereas baselines often degrade in specific task.
    }
    \label{tab:id_results}
    \renewcommand{\arraystretch}{0.8}
    {\scriptsize
    \begin{tabular}{lccccccccc}
    \toprule
    \multirow{2}{*}{Method} & \multirow{2}{*}{\stack{Codebook}{Size}} & \multicolumn{4}{c}{\textbf{Image (COCO)}}    &         & \multicolumn{3}{c}{\textbf{Audio (LibriSpeech)}} \\
    \cmidrule{3-6} \cmidrule{8-10}
    {}              & {}    & CVU    & LPIPS ↓      & PSNR ↑        & SSIM ↑       & & CVU    & PESQ ↑       & STOI ↑       \\
    \midrule
    VQ-VAE          & 256   & 0.3679 & 0.2053       & 23.1662       & 0.6704       & & -      & -            & -            \\
    SimVQ           & 256   & 0.9429 & \fst{0.1846} & \snd{23.5356} & \snd{0.6918} & & 0.0040 & 1.2524       & 0.4112       \\
    TokenBridge     & 256   & 0.9888 & 0.6075       & 11.5386       & 0.2026       & & 0.9888 & 1.1802       & 0.2881       \\
    FSQ             & 256   & 0.8607 & 0.2101       & 22.9390       & 0.6689       & & 0.0960 & 1.2763       & 0.7973       \\
    \textbf{FSP}    & 256   & 0.8852 & \snd{0.1884} & \fst{23.8131} & \fst{0.6940} & & 0.1372 & \snd{2.2653} & \snd{0.9083} \\
    \textbf{VP-VAE} & 256   & 0.7970 & 0.1929       & 23.4973       & 0.6881       & & 0.1352 & \fst{2.2718} & \fst{0.9100} \\
    \midrule
    VQ-VAE          & 1024  & 0.3557 & 0.1821       & 23.727        & 0.6946       & & 0.0010 & 1.4005       & 0.3628       \\
    SimVQ           & 1024  & 0.8919 & \fst{0.1662} & \snd{24.0102} & 0.7176       & & 0.0010 & 1.3213       & 0.4025       \\
    TokenBridge     & 1024  & 0.9887 & 0.6078       & 11.5145       & 0.1984       & & 0.9888 & 1.1801       & 0.3124       \\
    FSQ             & 1024  & 0.8059 & 0.1839       & 23.6006       & 0.6955       & & 0.0513 & 2.1738       & 0.9001       \\
    \textbf{FSP}    & 1024  & 0.8515 & 0.1722       & \fst{24.1910} & \snd{0.7177} & & 0.0810 & \fst{2.4458} & \fst{0.9191} \\
    \textbf{VP-VAE} & 1024  & 0.8102 & \snd{0.1717} & 23.8878       & \fst{0.7182} & & 0.0687 & \snd{2.3826} & \snd{0.9160} \\
    \midrule
    VQ-VAE          & 4096  & 0.2401 & 0.1693       & 24.1303       & 0.7245       & & -      & -            & -            \\
    SimVQ           & 4096  & 0.8604 & \fst{0.1497} & \fst{24.8338} & \snd{0.7455} & & 0.0002 & 1.2519       & 0.3972       \\
    TokenBridge     & 4096  & 0.9887 & 0.6086       & 11.5931       & 0.2011       & & 0.9887 & 1.2249       & 0.3160       \\
    FSQ             & 4096  & 0.7967 & 0.1647       & 24.4481       & 0.7273       & & 0.0294 & 2.3821       & 0.9182       \\
    \textbf{FSP}    & 4096  & 0.8227 & \snd{0.1521} & \snd{24.7291} & 0.7441       & & 0.0368 & \fst{2.4712} & \fst{0.9273} \\
    \textbf{VP-VAE} & 4096  & 0.8180 & 0.1587       & 24.6702       & \fst{0.7488} & & 0.0564 & \snd{2.4499} & \snd{0.9259} \\
    \midrule
    VQ-VAE          & 16384 & 0.2826 & 0.1489       & 24.5994       & 0.7555       & & -      & -            & -            \\
    SimVQ           & 16384 & 0.8276 & \fst{0.1387} & \fst{25.2429} & \snd{0.7605} & & 0.0001 & 1.2477       & 0.4436       \\
    TokenBridge     & 16384 & 0.9887 & 0.6143       & 11.4920       & 0.1963       & & 0.9887 & 1.2328       & 0.3008       \\
    FSQ             & 16384 & 0.7921 & 0.1552       & 24.5842       & 0.7354       & & 0.0149 & 2.4356       & 0.9225       \\
    \textbf{FSP}    & 16384 & 0.7115 & 0.1464       & 25.1676       & 0.7577       & & 0.0205 & \snd{2.5699} & \snd{0.9324} \\
    \textbf{VP-VAE} & 16384 & 0.7957 & \snd{0.1434} & \snd{25.2032} & \fst{0.7623} & & 0.0455 & \fst{2.5758} & \fst{0.9339} \\
    \midrule
\end{tabular}

    }
    \vspace{-0.18in}
\end{table*}

\begin{table*}[ht]
    \centering
    \caption{
        \textbf{Out-of-distribution generalization results.}
        Models trained on COCO/LibriSpeech are evaluated on unseen datasets: ImageNet (image) and Common Voice (audio).
        Our decoupled training paradigm yields superior generalization compared to baseline methods.
    }
    \label{tab:ood_results}
    \renewcommand{\arraystretch}{0.8}
    {\scriptsize
    \begin{tabular}{lccccccccc}
    \toprule
    \multirow{2}{*}{Method} & \multirow{2}{*}{\stack{Codebook}{Size}} & \multicolumn{4}{c}{\textbf{Image (ImageNet)}} & & \multicolumn{3}{c}{\textbf{Audio (Common Voice)}} \\
    \cmidrule{3-6} \cmidrule{8-10}
    {}              & {}    & CVU    & LPIPS ↓      & PSNR ↑         & SSIM ↑       & & CVU    & PESQ ↑       & STOI ↑       \\
    \midrule
    VQ-VAE          & 256   & 0.3617 & 0.2121       & 23.5942        & 0.6704       & & -      & -            & -            \\
    SimVQ           & 256   & 0.9363 & \fst{0.1927} & 23.6599        & \fst{0.6827} & & 0.0041 & 1.2103       & 0.3781       \\
    TokenBridge     & 256   & 0.9887 & 0.6079       & 11.6525        & 0.2024       & & 0.9887 & 1.0967       & 0.2688       \\
    FSQ             & 256   & 0.8574 & 0.2182       & 23.4555        & 0.6645       & & 0.1690 & 1.2064       & 0.7630       \\
    \textbf{FSP}    & 256   & 0.8819 & \snd{0.1989} & \snd{23.7919}  & 0.6736       & & 0.2860 & \snd{1.8148} & \snd{0.8537} \\
    \textbf{VP-VAE} & 256   & 0.7859 & \snd{0.1989} & \fst{23.8329}  & \snd{0.6759} & & 0.2055 & \fst{1.8510} & \fst{0.8617} \\
    \midrule
    VQ-VAE          & 1024  & 0.3604 & 0.1911       & 23.8520        & 0.6851       & & 0.0010 & 1.2103       & 0.3371       \\
    SimVQ           & 1024  & 0.8736 & \fst{0.1719} & \snd{24.3343}  & 0.7070       & & 0.0010 & 1.1568       & 0.3627       \\
    TokenBridge     & 1024  & 0.9887 & 0.6108       & 11.4561        & 0.1985       & & 0.9887 & 1.1059       & 0.2856       \\
    FSQ             & 1024  & 0.8000 & 0.1931       & 24.0973        & 0.6878       & & 0.1012 & 1.7776       & 0.8498       \\
    \textbf{FSP}    & 1024  & 0.8466 & 0.1795       & 24.2681        & \fst{0.7115} & & 0.1776 & \snd{1.9549} & \snd{0.8676} \\
    \textbf{VP-VAE} & 1024  & 0.7912 & \snd{0.1790} & \fst{24.4100}  & \fst{0.7115} & & 0.1441 & \fst{1.9697} & \fst{0.8701} \\
    \midrule
    VQ-VAE          & 4096  & 0.2371 & 0.1770       & 24.3765        & 0.7113       & & -      & -            & -            \\
    SimVQ           & 4096  & 0.8504 & \fst{0.1578} & 24.8154        & \snd{0.7339} & & 0.0002 & 1.1440       & 0.3713       \\
    TokenBridge     & 4096  & 0.9887 & 0.6071       & 11.5554        & 0.1920       & & 0.9888 & 1.1017       & 0.2787       \\
    FSQ             & 4096  & 0.7759 & 0.1749       & 24.4162        & 0.7065       & & 0.0670 & 1.9712       & 0.8707       \\
    \textbf{FSP}    & 4096  & 0.8074 & \snd{0.1598} & \snd{24.9106}  & 0.7323       & & 0.0936 & \snd{2.0564} & \snd{0.8840} \\
    \textbf{VP-VAE} & 4096  & 0.8004 & 0.1653       & \fst{24.9336}  & \fst{0.7400} & & 0.1151 & \fst{2.0841} & \fst{0.8907} \\
    \midrule
    VQ-VAE          & 16384 & 0.2621 & 0.1592       & 24.8349        & 0.7359       & & -      & -            & -            \\
    SimVQ           & 16384 & 0.8164 & \fst{0.1458} & 25.3304        & \snd{0.7496} & & 0.0001 & 1.1807       & 0.4002       \\
    TokenBridge     & 16384 & 0.9888 & 0.6166       & 11.4447        & 0.1969       & & 0.9887 & 1.1202       & 0.2722       \\
    FSQ             & 16384 & 0.7756 & 0.1627       & 24.8198        & 0.7237       & & 0.0499 & 1.9566       & 0.8721       \\
    \textbf{FSP}    & 16384 & 0.6887 & 0.1534       & \snd{25.3974}  & 0.7466       & & 0.0541 & \snd{2.1380} & \snd{0.8906} \\
    \textbf{VP-VAE} & 16384 & 0.7718 & \snd{0.1481} & \fst{25.4315 } & \fst{0.7535} & & 0.1218 & \fst{2.1803} & \fst{0.9003} \\
    \midrule
\end{tabular}

    }
    \vspace{-0.18in}
\end{table*}

\section{Experiments}\label{sec:exp}

\subsection{Experimental Setup}\label{subsec:exp_setup}

We evaluate VP-VAE and FSP on two tasks: image reconstruction and audio compression.
To comprehensively assess the trade-off between reconstruction fidelity and codebook utilization, we conduct experiments across four codebook sizes: \(K \in \{256, 1024, 4096, 16384\}\).

\subsubsection{Datasets and Backbones}
\textbf{Image.}
We use the \textbf{COCO 2017}~\cite{COCO2014} dataset for training.
Evaluation is performed on both the COCO validation set and the \textbf{ImageNet}~\cite{ImageNet2009} validation set to test out-of-distribution generalization.
All images are resized so that the shortest side is 256, then center-cropped to \(256 \times 256\).
For image experiments, we utilize the VQGAN architecture~\cite{esser2021}.
The encoder downsamples images by a factor of 8, resulting in a \(32 \times 32\) token grid for \(256^2\) inputs.

\textbf{Audio.}
Training is performed on the \textbf{LibriSpeech}~\cite{Librispeech2015} \texttt{train-clean-460} and \texttt{train-other-500} splits.
Evaluation uses the LibriSpeech \texttt{test-clean} and \texttt{test-other} splits, as well as the \textbf{Common Voice} (v18.0)~\cite{commonvoice} English test set.
All audio is resampled to 16kHz.
For audio experiments, we employ the encoder--decoder from SQCodec~\cite{SQCodec2025X}.
The temporal downsampling results in a frame rate of approximately 166.67 Hz.

To ensure a fair comparison that focuses solely on the quantization mechanism—and separates reconstruction capability from the hallucination effects of adversarial training—we do not use discriminators or GAN losses for any method.

\subsubsection{Evaluation Metrics}

\textbf{Reconstruction quality.}
We report standard fidelity metrics.
For images: \textbf{PSNR}, \textbf{SSIM}~\cite{SSIM2004}, and \textbf{LPIPS} (VGG)~\cite{LPIPS2018}.
For audio: \textbf{PESQ} (Wideband)~\cite{pesq2001} and \textbf{STOI}~\cite{stoi2010}.

\textbf{Codebook utilization.}
Standard ``codebook usage'' metrics, such as the percentage of codes used at least once, often saturate near 100\% on large validation sets and fail to reflect how \emph{evenly} the codebook is utilized.
To measure the balance of codebook utilization, we propose the \textbf{Codebook Valid Usage (CVU)}:
\begin{equation}
    \text{CVU} = \frac{\exp\left(-\sum_{c \in \mathcal{Q}} p(c) \log p(c)\right)}{K},
    \label{eq:cvu}
\end{equation}
where \(p(c)\) is the empirical selection probability of code \(c\) over all test tokens.
CVU represents the effective number of active codes divided by the target codebook size \(K\).
A CVU of 1.0 indicates a perfectly uniform utilization (maximum entropy), while lower values indicate imbalance.

\subsection{Results}\label{subsec:main_results}

We compare VP-VAE and FSP against the standard coupled baseline (VQ-VAE~\cite{VQVAE2017}), a state-of-the-art coupled method (SimVQ~\cite{SimVQ2025}), a state-of-the-art fixed-quantizer approach (FSQ~\cite{symmetry_FSQ2024}), and a recent predefined-grid discretization approach (TokenBridge~\cite{TokenBridge2025}).
Quantitative results for in-domain and out-of-distribution settings are presented in \cref{tab:id_results} and \cref{tab:ood_results}, respectively.

\textbf{Cross-Modality Consistency and Stability.}
A key finding is that VP-VAE and FSP exhibit superior adaptability across data modalities, whereas baselines tend to specialize or fail.
As shown in \cref{tab:id_results}, SimVQ excels on images, achieving the lowest LPIPS at all codebook sizes.
However, this performance does not transfer to audio.
On LibriSpeech, SimVQ exhibits near-collapse behavior in code usage, with extremely low CVU (from 0.004 at \(K{=}256\) down to \(10^{-4}\) at \(K{=}16384\)), accompanied by substantially degraded reconstruction quality (PESQ around 1.25--1.40), barely surpassing the classical VQ-VAE.
We hypothesize that this instability in audio stems from the statistical nature of the signal: silence and low-amplitude segments yield highly concentrated encoder outputs early in training, and coupled codebook learning can amplify small early assignment biases into a self-reinforcing imbalance (cf.\ \secref{sec:intro}) and leading to severe imbalance (or even training failure, as observed for VQ-VAE under some settings).

Conversely, FSQ is consistently stable on audio (e.g., STOI up to 0.9225 at \(K{=}16384\)) and avoids outright collapse due to its fixed quantization structure.
Nevertheless, its rigid grid prior can be suboptimal for image latents, leading to weaker reconstruction on COCO compared to our methods (e.g., at \(K{=}1024\), FSQ achieves 23.60 dB PSNR vs.\ 24.19 for FSP).

In contrast, VP-VAE and FSP deliver competitive or state-of-the-art results on \emph{both} modalities: on COCO, FSP achieves the highest PSNR at \(K{\in}\{256, 1024\}\), while VP-VAE obtains the best SSIM at \(K{\in}\{1024, 4096, 16384\}\); on LibriSpeech, both methods cconsistently occupy the top two positions in PESQ and STOI across all codebook sizes, substantially outperforming coupled baselines and improving over fixed quantization.
Overall, these results indicate that decoupling representation learning from discrete code optimization yields a more robust quantization mechanism that transfers across modalities without requiring modality-specific heuristic techniques.

\textbf{Out-of-Distribution Generalization.}
\cref{tab:ood_results} highlights the robustness of our approach under distribution shift.
On ImageNet, VP-VAE achieves the highest PSNR across all codebook sizes and the best SSIM at three of four cases.
The advantage is even more pronounced in audio.
On Common Voice, VP-VAE consistently achieves the best PESQ and STOI.
Notably, the gap between VP-VAE and strong baselines increases under OOD evaluation: for example, at \(K=16384\), VP-VAE's STOI score drops only marginally from \textbf{0.9339} (LibriSpeech) to \textbf{0.9003} (Common Voice).
In contrast, FSQ experiences a much sharper degradation, dropping from 0.9225 to 0.8721.

We attribute this robustness to the training objective: VP-VAE optimizes the decoder to tolerate a distribution of local perturbations rather than over-fitting to specific, singular codebook vectors.
This effectively regularizes the latent space, ensuring that quantization errors during inference (even if slightly shifted due to out-of-distribution inputs) remain within the decoder's tolerance.

\textbf{FSP v.s. FSQ.}
Our theoretical derivation in \secref{subsec:fsp} suggests that FSP is the theoretically principled generalization of FSQ for uniform latents.
The empirical data strongly supports this.
Across all experimental conditions—both modalities, all codebook sizes, and both ID/OOD evaluations—FSP consistently outperforms FSQ.
This validates that using Lloyd--Max centroids and proper perturbation intervals is superior to the rounding-to-integer heuristic employed by FSQ.
Crucially, FSP retains the computational simplicity of FSQ while closing the performance gap with fully adaptive methods like VP-VAE.

More experiments and analysis are provided in \secref{sec:other-experiments}.

    \section{Conclusion}\label{sec:conclusion}

We propose VP-VAE, a new perspective on discrete representation learning that treats quantization primarily as a \emph{structured latent perturbation} and, crucially, removes the need for an explicit codebook during training.
VP-VAE demonstrates that designing training-time perturbations to faithfully emulate inference-time quantization error is a principled and effective alternative to coupled codebook learning, offering a promising path toward stable, scalable discrete tokenizers for modern generative modeling.
Experiments on image and audio benchmarks validate the effectiveness of our approach.



    \newpage
\section*{Impact Statement}


This paper presents work whose goal is to advance the field of Machine
Learning. There are many potential societal consequences of our work, none
which we feel must be specifically highlighted here.



    \bibliography{refs}
    \bibliographystyle{icml2026}

\newpage
\appendix
\onecolumn

\section{Experimental Details}

\subsection{Representative Baselines}
We evaluate our approaches against four representative baselines:
\begin{enumerate}
    \item \textbf{VQ-VAE}~\cite{VQVAE2017}: The foundational coupled quantization baseline trained with the Straight-Through Estimator (STE).
    \item \textbf{SimVQ}~\cite{SimVQ2025}: A state-of-the-art coupled method that reparameterizes the codebook via a learnable linear layer to improve gradient flow.
    \item \textbf{FSQ}~\cite{FSQ2023}: A fixed scalar quantizer.
    We utilize the symmetric variant with noise injection~\cite{symmetry_FSQ2024, noise_FSQ2024} as a stronger baseline.
    The dimensionality \(d\) and the number of quantization levels per dimension are set according to the recommendations in~\cite{FSQ2023} to match the target codebook sizes \(K\):
    \begin{itemize}
        \item \(K=256\): Levels \([8, 6, 5]\) (\(d=3\)).
        \item \(K=1024\): Levels \([8, 5, 5, 5]\) (\(d=4\)).
        \item \(K=4096\): Levels \([7, 5, 5, 5, 5]\) (\(d=5\)).
        \item \(K=16384\): Levels \([8, 8, 8, 6, 5]\) (\(d=5\)).
    \end{itemize}
    \item \textbf{TokenBridge}~\cite{TokenBridge2025}: A method that discretizes the latent space using fixed Gaussian percentile grids.
    Following the official implementation, we first train a KL-regularized VAE to enforce a standard normal prior on the latent space.
    Subsequently, we discretize the continuous representations using a fixed grid derived from the percentiles of the Gaussian distribution (4 bins per dimension).
\end{enumerate}

For FSP implementation, we employ the exact same dimensionality \(d\) and level configurations (\([L_1, \dots, L_d]\)) as FSQ for each target codebook size.
We utilize the Tanh-based bounded activation, consistent with FSQ.
For VP-VAE, we also set the bottleneck dimension \(d\) (cf. Eq.~\ref{eq:proj}) to match the dimensionality of the corresponding FSQ/FSP configuration (e.g., \(d=3\) for \(K=256\)). This ensures that the density estimation complexity and the representational dimension remain consistent across baselines.

\subsection{Training and Optimization}

All models are trained for 50 epochs using the same optimizer with the same hyperparameter settings as in the original paper~\cite{esser2021, SQCodec2025X}.
For images, we minimize a combination of \(L_1\) loss and LPIPS perceptual loss~\cite{LPIPS2018} (VGG backbone).
For audio, we use an \(L_1\) reconstruction loss in the time domain, a multi-resolution STFT loss, and a perceptual loss based on intermediate features from a frozen Whisper encoder~\cite{Whisper2023}.
All experiments are conducted on 2 NVIDIA RTX4090 GPUs.

\section{Additional Experiments}\label{sec:other-experiments}

\subsection{Ablation Study}\label{subsec:ablation_study}

We conduct ablation studies to validate the key design choices in VP-VAE and FSP.
Unless otherwise specified, experiments are conducted on the image modality with a target codebook size of \(K{=}1024\).

\subsubsection{VP-VAE Ablation}

\begin{table}[h]
    \centering
    \caption{\textbf{Ablation on VP-VAE components.}
    We evaluate the contribution of latent normalization and Metropolis--Hastings mechanism on image reconstruction (\(K{=}1024\)).
    Both components contribute to reconstruction quality and codebook balance.}
    \label{tab:ablation_result}
    \begin{tabular}{lcccc}
        \toprule
        Setting                                 & CVU    & LPIPS ↓ & PSNR ↑  & SSIM ↑ \\
        \midrule
        VP-VAE (full)                           & 0.8102 & 0.1717  & 23.8878 & 0.7182 \\
        \quad w/o \(\mathcal{L}_{\text{norm}}\) & 0.8052 & 0.1850  & 24.0219 & 0.7167 \\
        \quad w/o MH (always accept)            & 0.7467 & 0.1756  & 23.8053 & 0.7138 \\
        \bottomrule
    \end{tabular}
\end{table}

\paragraph{Effect of latent normalization.}
The normalization regularizer (Eq.~\ref{eq:loss_norm}) is important for the stability of our scale estimation mechanism.
Since the kNN-based radius estimation (Eq.~\ref{eq:radius}) relies on Euclidean distance, it is sensitive to the relative scaling of latent dimensions.
Without normalization, dimensions with naturally larger variances can dominate the distance calculation, making the kNN-based radius estimation in Eq.~\eqref{eq:radius} less reliable and consequently weakening the scale alignment of perturbations.
As shown in \cref{tab:ablation_result}, removing \(\mathcal{L}_{\text{norm}}\) degrades perceptual reconstruction quality (LPIPS) and slightly reduces utilization balance.

\paragraph{Effect of Metropolis--Hastings mechanism.}
We compare our full MH procedure against a simplified variant that always accepts proposals (equivalent to injecting uniform noise within the estimated radius).
The MH step acts as a filter for \emph{distribution consistency}: it rejects perturbations that would push the latent vector into low-density regions or out-of-distribution areas.
Without MH, the decoder is forced to reconstruct from ``invalid latent states'', effectively wasting model capacity.
\cref{tab:ablation_result} confirms that removing the acceptance criterion reduces both CVU (0.81 \(\to\) 0.75) and reconstruction quality.

\subsubsection{FSP Ablation}

\begin{table}[h]
    \centering
    \caption{\textbf{FSP activation functions on images.}
    We compare three CDF-like activations, which affect how well the ``approximately uniform'' assumption holds and thus the effectiveness of FSP (\(K{=}1024\)).
    }
    \label{tab:fsp_act_image}
    \begin{tabular}{lcccc}
        \toprule
        Setting           & CVU    & LPIPS ↓ & PSNR ↑  & SSIM ↑ \\
        \midrule
        FSP (Tanh)        & 0.8515 & 0.1722  & 24.1910 & 0.7177 \\
        FSP (Normal CDF)  & 0.8663 & 0.1689  & 24.2945 & 0.7254 \\
        FSP (Laplace CDF) & 0.8390 & 0.1685  & 24.2433 & 0.7263 \\
        \bottomrule
    \end{tabular}
\end{table}

\begin{table}[h]
    \centering
    \caption{\textbf{FSP activation functions on audio.}
    We evaluate three CDF-like activations on audio (\(K{=}1024\)).
    }
    \label{tab:fsp_act_audio}
    \begin{tabular}{lccc}
        \toprule
        Setting           & CVU    & PESQ ↑ & STOI ↑ \\
        \midrule
        FSP (Tanh)        & 0.0810 & 2.4458 & 0.9191 \\
        FSP (Normal CDF)  & 0.0654 & 2.3772 & 0.9154 \\
        FSP (Laplace CDF) & 0.0751 & 2.3760 & 0.9198 \\
        \bottomrule
    \end{tabular}
\end{table}

\paragraph{Choice of FSP activation.}
FSP assumes the latent variables can be mapped to an approximate uniform distribution via an activation function \(g(\cdot)\).
We compare three choices: Tanh (rescaled), Normal CDF, and Laplace CDF.
As shown in \cref{tab:fsp_act_image}, different activations yield modest trade-offs across metrics.
We also evaluate these activations on the audio modality (\cref{tab:fsp_act_audio}) and observe similar patterns: no single activation dominates across all metrics, suggesting that the choice can be tuned per application.

\subsection{Qualitative Analysis}\label{subsec:qualitative}

\begin{figure}[h]
    \centering
    \includegraphics[width=0.6\linewidth]{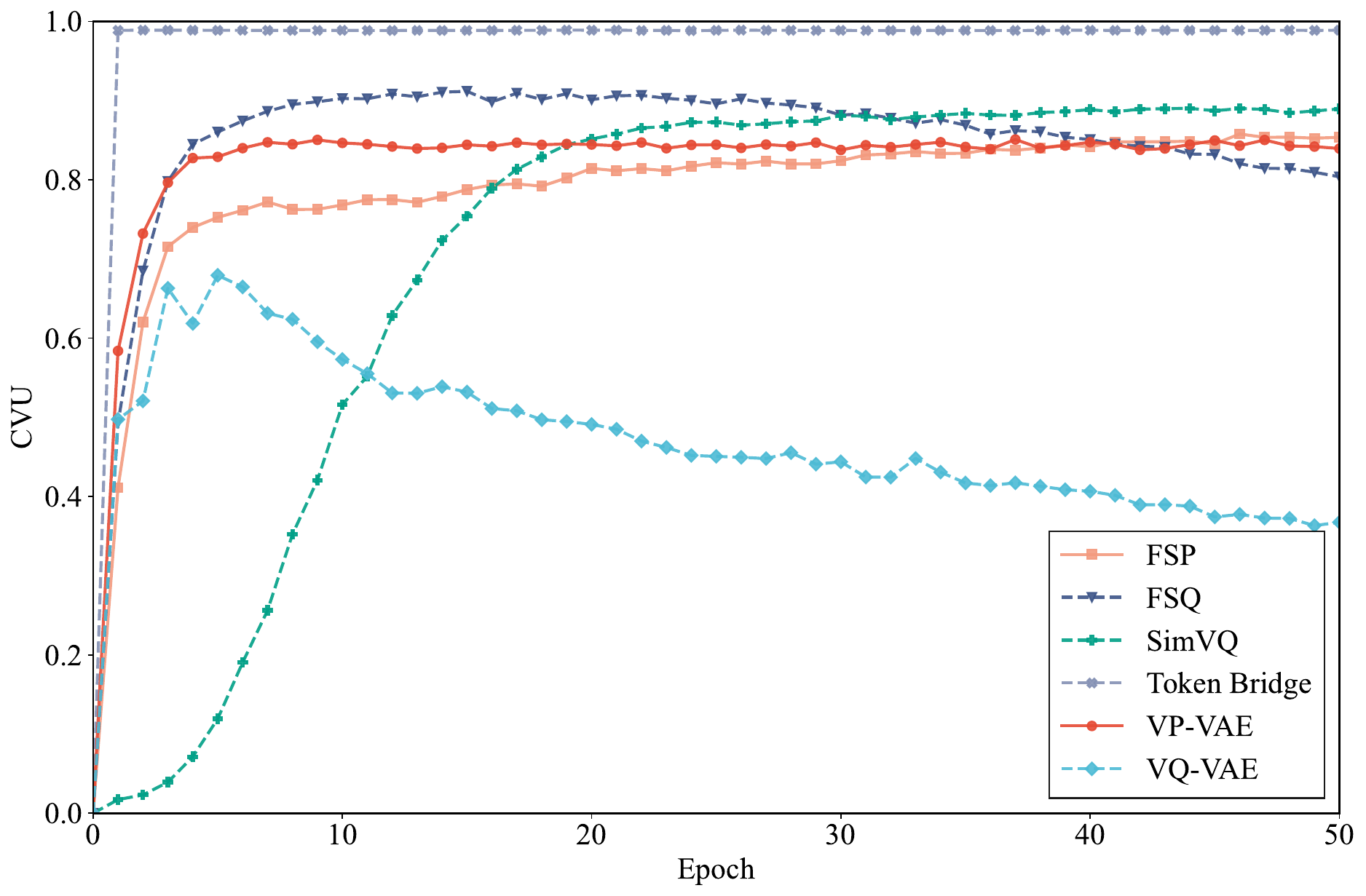}
    \caption{\textbf{Codebook utilization during training.}
    CVU curves for different methods on image reconstruction (\(K{=}1024\)).
    VQ-VAE and FSQ exhibit an initial rise followed by a decline. VP-VAE and FSP maintain stable, high utilization throughout training.}
    \label{fig:cvu_curve}
    \vspace{-0.1in}
\end{figure}

\paragraph{Training dynamics of codebook utilization.}
\cref{fig:cvu_curve} visualizes CVU over training epochs for different methods at \(K{=}1024\).
We observe two distinct behaviors regarding stability.
First, VQ-VAE and FSQ exhibit an initial rise followed by a decline: utilization increases initially as the encoder explores the space, but subsequently degrades.
This decline reflects certain codes become increasingly dominant, triggering the self-reinforcing vicious cycle where rarely used codes receive fewer gradient updates and become even less likely to be chosen.(The trend of FSQ is slower than VQ-VAE due to the fixed codebook.)
In contrast, VP-VAE and FSP maintain stable CVU throughout training, demonstrating the stability benefits of decoupled training.

Second, we analyze the relationship between utilization and quality.
TokenBridge achieves near-perfect utilization (CVU \(\approx 0.99\)) yet yields significantly lower reconstruction quality (e.g., LPIPS \(\approx 0.6\)).
This phenomenon can be explained by the trade-off between capacity and granularity.
The codebook size determines the discrete representation capacity and directly controls the granularity of quantization: larger codebooks yield finer partitions, while smaller codebooks impose coarser discretization.
TokenBridge is originally designed for massive codebooks (\(K \geq 2^{48}\)) where a fixed grid provides sufficient resolution.
At the moderate codebook sizes evaluated here (\(K \leq 16384\)), the rigid Gaussian percentile grid is too coarse to capture fine-grained data details, regardless of how uniformly the bins are used.
This underscores that high CVU alone is insufficient; effective discretization requires that the induced quantization error be compatible with the decoder's robustness.

Consistent with the image results in \cref{tab:id_results}, only SimVQ, VP-VAE, and FSP achieve a favorable balance of both high utilization and high fidelity in the image domain.

\begin{figure}[h]
    \centering
    \includegraphics[width=0.5\linewidth]{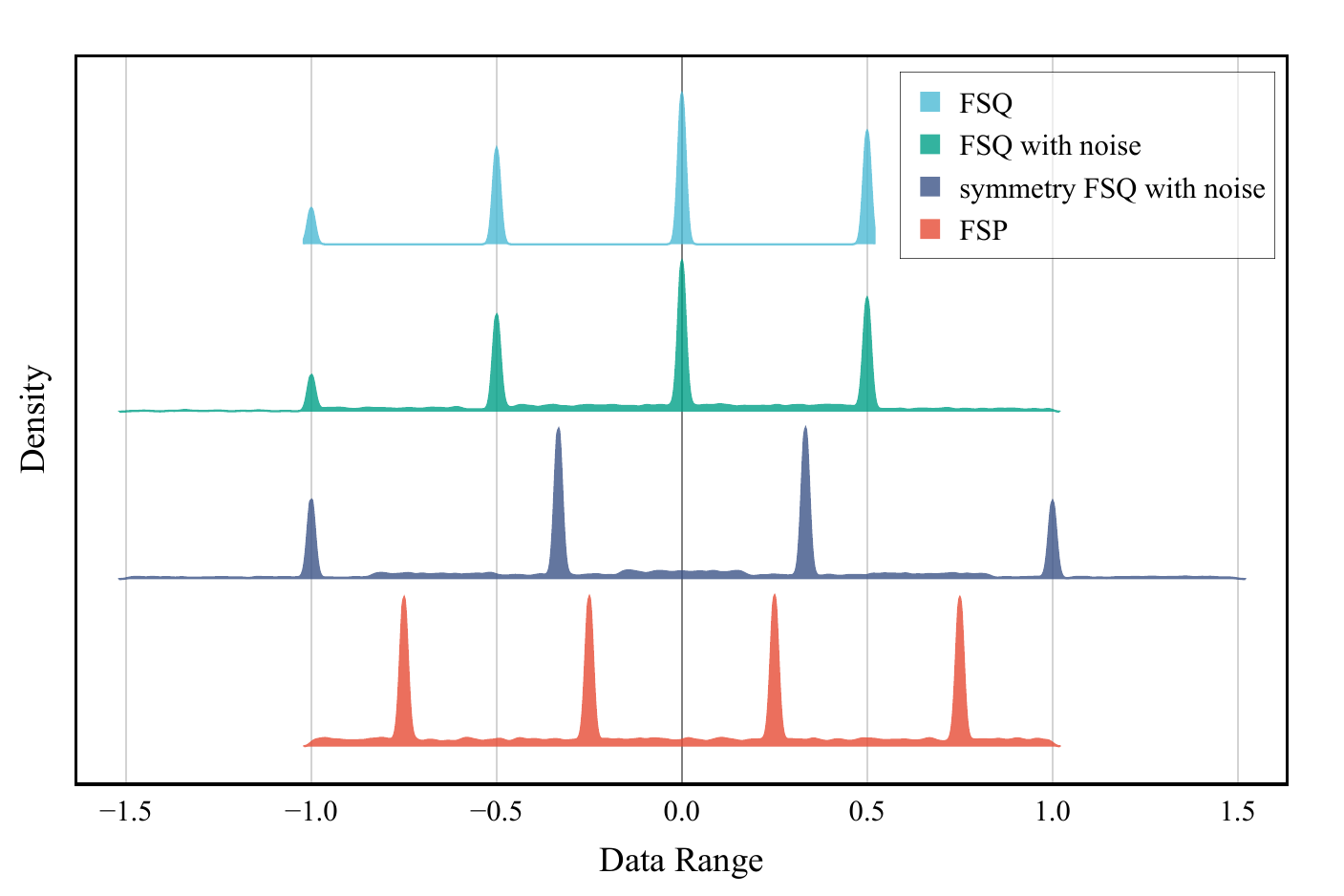}
    \caption{\textbf{Output distributions of fixed quantization schemes.} Given a uniform latent distribution, we compare the quantized output distributions produced by FSQ, FSQ with noise, symmetric FSQ with noise, and FSP. They are all configured with \(L{=}4\) quantization levels.
    FSP produces a more uniform output distribution, aligning with the Lloyd--Max optimality principle.}
    \label{fig:fsp_vs_fsq}
    \vspace{-0.15in}
\end{figure}

\paragraph{Comparison of FSP and FSQ quantization schemes.}
\cref{fig:fsp_vs_fsq} provides an intuitive comparison of how different fixed quantization schemes transform a uniform latent distribution.
With \(L{=}4\) quantization levels, standard FSQ rounds to grid boundaries, producing a non-uniform output distribution skewed toward the boundaries.
In contrast, FSP quantizes to interval centroids, which are precisely the Lloyd--Max optimal reconstruction points for a uniform source.
This produces an output distribution that remains approximately uniform, explaining FSP's consistent advantage over FSQ variants in our experiments.

\section{Limitations and Future Work}\label{sec:limitations.}

While VP-VAE offers stable and high-fidelity training, it relies on non-parametric density estimation via \(k\)-nearest-neighbor search, which introduces additional computational overhead during training, particularly as the size of the memory queue \(\mathcal{S}\) grows.
We mitigate this cost through random subsampling and by applying perturbations within a low-dimensional bottleneck. 
We envision that further acceleration techniques (e.g., approximate nearest-neighbor search) could be beneficial for scaling to very large models.
Additionally, our current experiments focus on autoencoder-based tokenization; integrating VP-VAE with autoregressive or diffusion-based token generators remains an interesting direction.

\end{document}